\documentclass{article}
\usepackage[preprint]{colm2026_conference}
\usepackage{microtype}
\usepackage[hyphens]{url}
\usepackage{hyperref}
\usepackage{booktabs}
\usepackage{amsmath,amssymb,amsfonts}
\usepackage{graphicx}
\usepackage{algorithm}
\usepackage{algpseudocode}

\definecolor{darkblue}{rgb}{0, 0, 0.5}
\hypersetup{colorlinks=true, citecolor=darkblue, linkcolor=darkblue, urlcolor=darkblue}

\title{Cited but Not Verified: Parsing and Evaluating \\Source Attribution in LLM Deep Research Agents}

\author{\textbf{Hailey Onweller}\textsuperscript{*} \quad
\textbf{Elias Lumer} \quad
\textbf{Austin Huber} \\
\textbf{Pia Ramchandani} \quad
\textbf{Vamse Kumar Subbiah} \quad
\textbf{Corey Feld} \\
Commercial Technology and Innovation Office, PricewaterhouseCoopers, U.S.
}

\begin{document}

\maketitle

\begin{abstract}
Large language models (LLMs) power deep research agents that synthesize information from hundreds of web sources into cited reports, yet these citations cannot be reliably verified. Current approaches either trust models to self-cite accurately, risking bias, or employ retrieval-augmented generation (RAG) that does not validate source accessibility, relevance, or factual consistency. We introduce the first source attribution evaluation framework that uses a reproducible AST parser to extract and evaluate inline citations from LLM-generated Markdown reports at scale. Unlike methods that verify claims in isolation, our framework closes the loop by retrieving the actual cited content, enabling human or model evaluators to judge each citation against its source. Citations are evaluated along three dimensions. (1) \textit{Link Works} verifies URL accessibility, (2) \textit{Relevant Content} measures topical alignment, and (3) \textit{Fact Check} validates factual accuracy against source content. We benchmark 14 closed-source and open-source LLMs across three evaluation dimensions using rubric-based LLM-as-a-judge evaluators calibrated through human review. Our results reveal that even the strongest frontier models maintain link validity above 94\% and relevance above 80\%, yet achieve only 39--77\% factual accuracy, while fewer than half of open-source models successfully generate cited reports in a one-shot setting. Ablation studies on research depth show that Fact Check accuracy drops by approximately 42\% on average across two frontier models as tool calls scale from 2 to 150, demonstrating that more retrieval does not produce more accurate citations. These findings reveal a critical disconnect between surface-level citation quality and factual reliability, and our framework provides the evaluation infrastructure to assess the disconnect.
\end{abstract}

\section{Introduction}

Deep research agents powered by Large Language Models (LLMs) now synthesize information from hundreds of web sources into comprehensive reports with inline citations, enabling systems such as Perplexity AI, ChatGPT with web search, and Google Gemini to promise verifiable research at scale~\citep{perplexity2024, openai2024search}. However, the citations these agents produce cannot be reliably verified. Current approaches either trust models to self-cite accurately, risking hallucinated or misattributed references~\citep{halogen2024}, or employ retrieval-augmented generation (RAG) that does not validate whether cited sources are accessible, topically relevant, or factually consistent with the claims they support~\citep{lewis2020rag, lumer2025comparison, gulati2026rows}. Recent studies have documented citation hallucination rates ranging from 11\% to 57\% across commercially deployed models~\citep{yuan2026citeaudit}, yet no existing framework evaluates citation quality beyond binary attribution verification. This gap between deployment scale and evaluation rigor raises a critical question. \textit{How reliable are the citations that millions of users encounter daily in LLM-generated research?}

Despite growing recognition of this problem, existing evaluation approaches remain insufficient. Benchmarks such as AttributionBench~\citep{attributionbench2024}, CiteME~\citep{citeme2024}, and CiteEval~\citep{citeeval2025} focus on binary attribution classification, citation matching, or paradigm comparison, while CiteAudit~\citep{yuan2026citeaudit} targets fabricated references in scientific writing. Three critical gaps persist. (1) No end-to-end framework combines citation extraction with multi-dimensional quality assessment across URL accessibility, topical relevance, and factual accuracy, (2) no systematic comparison exists across major LLM providers in deep research settings, and (3) the relationship between search depth and citation quality remains unexplored. These gaps motivate the following research questions.

\begin{quote}
\textit{To what extent do frontier LLMs produce factually accurate citations, and does surface-level citation quality (link validity, topical relevance) mask deeper factual failures? How does increasing search depth affect citation quality across these dimensions?}
\end{quote}

\begin{figure}[t]
\centering
\includegraphics[width=\columnwidth]{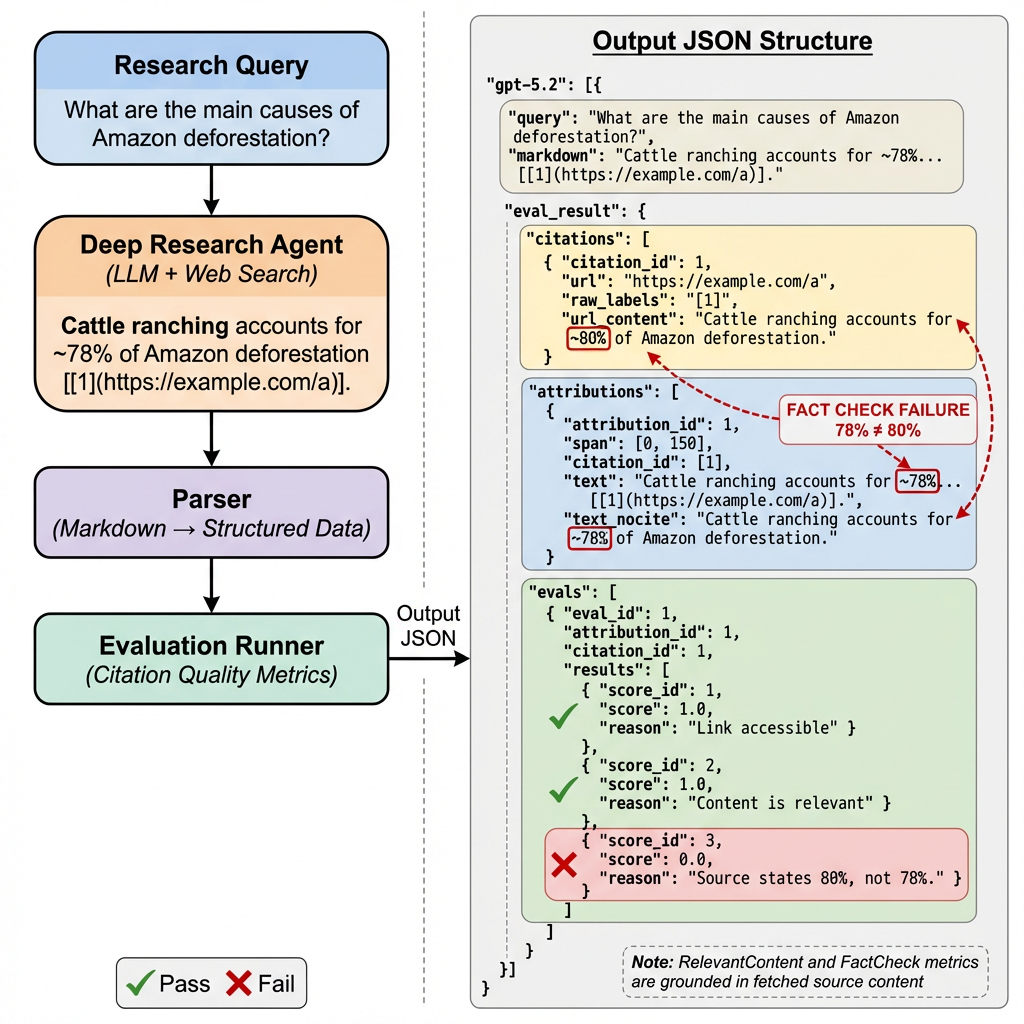}
\caption{Source attribution evaluation framework. A deep research agent generates Markdown reports with inline citations, which are parsed via a Markdown AST parser to extract citation-claim pairs. Each pair is evaluated on Link Works (URL accessibility), Relevant Content (topical alignment), and Fact Check (factual accuracy).}
\label{fig:architecture}
\end{figure}

To address these questions, we introduce a source attribution evaluation framework that extracts and evaluates inline citations from LLM-generated Markdown reports at scale. As illustrated in Figure~\ref{fig:architecture}, the framework operates as a three-stage pipeline. First, a Markdown Abstract Syntax Tree (AST) parser structurally extracts citation-claim pairs without requiring LLM inference. Second, each cited URL is retrieved and its content extracted. Third, three complementary evaluators assess citation quality. (1) \textit{Link Works} verifies URL accessibility, (2) \textit{Relevant Content} measures topical alignment via LLM-as-a-judge, and (3) \textit{Fact Check} validates factual accuracy against retrieved source content via LLM-as-a-judge calibrated through human review.

We benchmark 14 LLMs spanning open-source and closed-source frontier models across diverse research queries. Our evaluation reveals three key findings. First, even the strongest frontier models maintain link validity above 94\% and content relevance above 80\%, yet achieve only 39--77\% factual accuracy, exposing a critical disconnect between surface-level citation quality and factual reliability. Second, fewer than half of open-source models successfully generate cited reports (17--40\% task success), compared to 83--100\% for frontier models. Third, our ablation study across two models demonstrates that Fact Check accuracy drops approximately 42\% on average as search depth scales from 2 to 150 tool calls, while Link Works and Relevant Content remain stable, suggesting that information overload impairs factual synthesis rather than improving it.

We make the following contributions:
\begin{itemize}
    \item We introduce the first end-to-end source attribution evaluation framework that combines deterministic citation extraction with multi-dimensional quality assessment across link accessibility, topical relevance, and factual accuracy.
    \item We benchmark 14 LLMs across three major providers and open-source alternatives, revealing that factual accuracy is the most challenging and differentiating dimension of citation quality.
    \item We demonstrate through ablation studies that increased search depth consistently degrades factual accuracy while surface-level citation metrics remain stable, providing evidence for an information overload effect in LLM research synthesis.
\end{itemize}

\section{Related work}

\subsection{Attributed generation and deep research systems}

LLMs can generate responses with inline citations through web search integration and retrieval-augmented generation (RAG)~\citep{lewis2020rag}. Commercial deep research systems such as Perplexity AI, ChatGPT with web search, and Google Gemini with grounding have made citation-generating LLMs widely accessible~\citep{perplexity2024, openai2024search, lumer2025toolagentsurvey, lumer2025tool}. Model providers have also introduced API-level citation support, such as Anthropic's Citations API, which returns structured references with character-level provenance to source documents~\citep{anthropic2025citations}. However, these features focus on citation \textit{generation} rather than citation \textit{evaluation}. Research on attributed generation has explored both generation-time approaches that produce citations during output~\citep{gao2023enabling, slobodkin2024attribute} and post-hoc methods that add citations after text generation~\citep{gao2023rarr}. \citet{wan2025generationprograms} introduce a modular framework that decomposes attributed generation into executable programs, achieving fine-grained sentence-level attribution. \citet{wang2025scholarcopilot} propose a unified generation-retrieval model for academic writing that dynamically triggers citation retrieval during text generation. Despite these advances, citation hallucination rates of 11--57\% persist across commercially deployed models~\citep{yuan2026citeaudit}, and citations may link to inaccessible URLs, reference irrelevant content, or misrepresent source claims~\citep{halogen2024}. Unlike these works, which focus on improving citation generation, our work focuses on evaluating the quality of citations already produced by deployed systems.

\subsection{Attribution evaluation benchmarks}

Several benchmarks have been developed for evaluating LLM-generated attributions. AttributionBench~\citep{attributionbench2024} provides a benchmark for automatic binary attribution classification, finding that even fine-tuned GPT-3.5 achieves only 80\% macro-F1. CiteME~\citep{citeme2024} evaluates citation matching in academic contexts, revealing that LLMs achieve only 4--18\% accuracy on identifying correct papers to cite. CiteEval~\citep{citeeval2025} goes beyond NLI-based approaches by conducting fine-grained citation assessment across full retrieval contexts. CiteGuard~\citep{citeguard2025} reframes citation evaluation as attribution alignment, achieving 68\% accuracy on CiteME through retrieval-aware verification. RefChecker~\citep{refchecker2024} presents a claim-triplet framework for fine-grained hallucination detection, while CiteAudit~\citep{yuan2026citeaudit} addresses fabricated references in scientific writing through multi-agent verification. Most recently, \citet{saxena2025gcite} compare generation-time versus post-hoc citation paradigms, finding a consistent trade-off between coverage and correctness. \citet{seo2025verifying} evaluate fact verifiers across 14 benchmarks and find that annotation ambiguity substantially affects model rankings, highlighting the importance of careful benchmark design for evaluation tasks. While these works advance attribution evaluation, they primarily address binary verification, citation matching, or paradigm comparison. Our framework differs by evaluating citations across three complementary quality dimensions simultaneously and by targeting live web citations in deep research contexts rather than controlled document sets.

\subsection{LLM-as-a-judge for evaluation}

The use of LLMs as automated evaluators has gained significant attention as an alternative to costly human annotation~\citep{zheng2024judging}. However, research has revealed systematic biases in LLM judges, including position bias, verbosity bias, and self-enhancement effects~\citep{wang2023large, ye2024justice}. \citet{wang2025judging} demonstrate that even Large Reasoning Models remain susceptible to evaluation biases despite advanced reasoning capabilities. These findings are directly relevant to our framework, which relies on LLM-as-a-judge evaluators for the Relevant Content and Fact Check dimensions. To mitigate potential judge biases, we calibrate our evaluators through human review and employ rubric-based scoring that constrains evaluation to specific factual criteria rather than open-ended quality assessment. Our work extends the LLM-as-a-judge paradigm from general text evaluation to the specific task of source attribution verification, where the evaluator must assess whether a claim is supported by retrieved source content rather than judging overall response quality.

\section{Methodology}

\subsection{Overview}

Our framework evaluates source attribution quality in LLM-generated research responses through a three-stage pipeline (Figure~\ref{fig:architecture}). First, a deep research agent generates a comprehensive Markdown report with inline citations for a given query. Second, a Markdown Abstract Syntax Tree (AST) parser structurally extracts citation-claim pairs without requiring LLM inference. Third, three complementary evaluators assess each citation along distinct quality dimensions. The pipeline processes attributions independently at the sentence level, enabling fine-grained analysis rather than document-level assessment. This modular design enables evaluation of any LLM capable of generating Markdown responses with citations, requires no modification to the underlying model, and scales to thousands of cited pages. In Algorithm~\ref{alg:pipeline}, the deep research agent's Markdown output is parsed into an \texttt{AttributionDocument} of citation-claim pairs, and the evaluation runner then fetches each cited source and scores every pair across the three dimensions.

\begin{algorithm}[t]
\caption{Source Attribution Evaluation Pipeline}\label{alg:pipeline}
\begin{algorithmic}[1]
\Require Research query $q$, deep research agent with web search
\Ensure \texttt{AttributionDocument} $\mathcal{D}$ with \texttt{citations}, \texttt{attributions}, \texttt{evals}
\Statex \textit{Phase 0: Report Generation}
\State $D \gets \Call{DeepResearchAgent}{q}$ \Comment{Markdown report with inline citations}
\Statex \textit{Phase 1: Markdown AST Parsing}
\State $D' \gets \Call{Canonicalize}{D}$ \Comment{Normalize whitespace, strip code blocks}
\State $T \gets \Call{BuildAST}{D'}$ \Comment{Construct abstract syntax tree}
\State $\mathcal{C} \gets \Call{ExtractCitations}{T}$ \Comment{Deduplicated \texttt{url}, \texttt{raw\_labels}}
\State $\mathcal{S} \gets \Call{SentenceSegment}{T}$ \Comment{Split into individual claims}
\State $\mathcal{A} \gets \Call{BackwardAttribute}{\mathcal{S}, \mathcal{C}}$ \Comment{\texttt{text\_nocite}, \texttt{span}, \texttt{citation\_ids}}
\State $\mathcal{D} \gets \texttt{AttributionDocument}(\texttt{citations}{=}\mathcal{C},\; \texttt{attributions}{=}\mathcal{A})$
\Statex \textit{Phase 2: Evaluation Runner}
\ForAll{unique citation $c \in \mathcal{C}$ \textbf{in parallel}}
    \State $c.\texttt{url\_content} \gets \Call{Fetch}{c.\texttt{url}}$ \Comment{Web content extraction}
\EndFor
\ForAll{attribution-citation pair $(a_i, c_j) \in \mathcal{D}$ \textbf{in parallel}}
    \State $s_1 \gets \Call{LinkWorks}{c_j.\texttt{url}}$ \Comment{HTTP accessibility}
    \State $s_2 \gets \Call{RelevantContent}{a_i.\texttt{text\_nocite},\; c_j.\texttt{url\_content}}$ \Comment{LLM-as-a-judge}
    \State $s_3 \gets \Call{FactCheck}{a_i.\texttt{text\_nocite},\; c_j.\texttt{url\_content}}$ \Comment{LLM-as-a-judge}
    \State $\mathcal{D}.\texttt{evals} \gets \mathcal{D}.\texttt{evals} \cup \{(a_i.\texttt{id},\; c_j.\texttt{id},\; [s_1, s_2, s_3])\}$
\EndFor
\State \Return $\mathcal{D}$
\end{algorithmic}
\end{algorithm}

\subsection{Markdown AST parser}

The parser transforms raw LLM-generated Markdown into structured citation-claim pairs by constructing an abstract syntax tree and traversing it to extract citation nodes. Given a Markdown document, it produces an \texttt{AttributionDocument} containing a deduplicated list of citations (URLs), a list of attributions (text spans with associated citation references), and placeholder structures for evaluation results.

Parsing proceeds in four stages. Canonicalization normalizes line endings and whitespace. Code block removal strips fenced code sections to prevent false citation matches. AST construction and traversal identifies citation nodes across multiple formats, including numbered references (\texttt{[1]}, \texttt{[2]}), footnote-style references (\texttt{[\^{}note]}), inline Markdown links (\texttt{[text](url)}), autolinks (\texttt{<url>}), and ranges (\texttt{[1-3]} expanding to individual references). Registry building creates a deduplicated citation list with normalized URLs.

The parser performs sentence-level segmentation to split continuous text into individual claims. It implements backward attribution logic. When a citation appears at the end of a passage, it applies to all preceding uncited sentences in that passage. This design reflects common LLM citation practices, where a single reference supports multiple related claims. The final output associates each text span with its citation identifiers, enabling independent evaluation of each claim-citation pair.

\subsection{Evaluation dimensions}

The framework evaluates citations across three complementary dimensions, each capturing a distinct aspect of attribution quality. These dimensions form an increasing hierarchy of verification difficulty.

\subsubsection{Link Works}

Link Works assesses URL accessibility without LLM inference. For each cited URL, a web content extractor capable of handling JavaScript-rendered pages retrieves the content. The evaluator produces a binary score of 1 if the URL returns accessible content and 0 if the request fails due to HTTP errors (404, 403), timeouts, or blocked access. This dimension identifies broken links, paywalled content, and URLs removed since the research was conducted.

\subsubsection{Relevant Content}

Relevant Content measures topical alignment between the claim and the cited source using an LLM-as-a-judge approach. Given the attribution text and retrieved source content (truncated to 5,000 characters), the evaluator determines whether the source addresses the same topic as the claim. The evaluator produces a binary score with a natural language explanation. This dimension identifies citations that link to valid URLs but reference content unrelated to the claim's subject matter.

\subsubsection{Fact Check}

Fact Check verifies whether specific factual claims are accurately supported by the source content. Using an LLM-as-a-judge approach, the evaluator examines facts, numbers, dates, and assertions in the attribution text against the retrieved source. The evaluator produces a binary score of 1 if the facts are supported or consistent and 0 if they are contradicted, absent, or uncertain. This dimension represents the most stringent evaluation, identifying citations where the source exists and is topically relevant but does not support the specific claims attributed to it. To ensure alignment with human judgment, the Fact Check evaluator was calibrated through manual review of 50--100 LLM judgments.

\subsection{Experimental setup}

\textbf{Models and queries.} We evaluate 14 LLMs spanning three major providers and open-source alternatives on 130 research queries drawn from DeepResearch Bench~\citep{du2025deepresearch} and BrowseComp~\citep{wei2025browsecomp}. The models include OpenAI (GPT-5.2, GPT-5.4, GPT-5 Mini, Codex), Anthropic (Claude Sonnet 4.5, Claude Sonnet 4.6, Claude Opus 4.5, Claude Opus 4.6, Claude Haiku 4.5), Google (Gemini 3.1 Pro, Gemini 3 Flash), and three open-source models (OSS-120B, Llama 4 Maverick, Pixtral Large). Queries cover diverse topics requiring multi-source synthesis.

\textbf{Evaluation protocol.} For each query, a deep research agent with web search capabilities generates a Markdown response with inline citations. The agent is configured with a system prompt enforcing citation format requirements and minimum search depth. Evaluation runs asynchronously with concurrency limits (10 concurrent agents, 15 concurrent evaluators) and retry logic (5 retries with 5-second delays) to handle transient failures.

\textbf{Ablation study.} To analyze the relationship between search depth and attribution quality, we vary the maximum number of tool calls across two models (GPT-5.4, Claude Opus 4.6) at seven intervals (2, 10, 30, 50, 70, 100, and 150). This controlled experiment isolates the effect of search depth from model capability differences.

\section{Evaluation}

We report results organized around three key findings that address our research questions. Section~\ref{sec:finding1} examines the disconnect between surface-level citation metrics and factual accuracy, Section~\ref{sec:finding2} analyzes provider-level patterns in citation quantity versus quality, and Section~\ref{sec:finding3} investigates the effect of search depth on attribution quality.

\subsection{Finding 1. Surface-level citation quality masks factual failures}
\label{sec:finding1}

Table~\ref{tab:main_results} presents evaluation results across all 14 models.

\begin{table}[t]
\centering
\caption{Source attribution quality across 14 LLMs ordered by Relevant Content. Success Rate indicates percentage of queries producing valid citations. \textbf{Bold} indicates best per metric.}
\label{tab:main_results}
\begin{tabular}{lcccc}
\toprule
\textbf{Model} & \textbf{Success} & \textbf{Link Works} & \textbf{Relevant} & \textbf{Fact Check} \\
\midrule
Claude Opus 4.5 & 90.0\% & 98.7\% & \textbf{95.7\%} & \textbf{76.8\%} \\
GPT-5.4 & \textbf{100.0\%} & \textbf{100.0\%} & 93.7\% & 47.7\% \\
GPT-5.2 & \textbf{100.0\%} & 98.3\% & 92.3\% & 58.8\% \\
Codex & \textbf{100.0\%} & 96.9\% & 91.9\% & 54.1\% \\
Claude Haiku 4.5 & 83.3\% & 98.9\% & 91.1\% & 68.9\% \\
Claude Sonnet 4.6 & 93.3\% & 99.2\% & 89.8\% & 58.7\% \\
Claude Sonnet 4.5 & 96.7\% & 98.9\% & 88.3\% & 51.8\% \\
GPT-5 Mini & \textbf{100.0\%} & 99.3\% & 87.4\% & 38.9\% \\
Claude Opus 4.6 & 93.3\% & 97.2\% & 83.9\% & 54.2\% \\
Gemini 3 Flash & \textbf{100.0\%} & 94.7\% & 82.9\% & 45.2\% \\
Gemini 3.1 Pro & 90.0\% & 94.1\% & 80.7\% & 48.5\% \\
OSS-120B & 40.0\% & 83.9\% & 68.7\% & 24.4\% \\
Pixtral Large & 16.7\% & \textbf{100.0\%} & 64.9\% & 51.4\% \\
Llama 4 Maverick & 30.0\% & 80.8\% & 60.6\% & 34.3\% \\
\bottomrule
\end{tabular}
\end{table}

High link validity and relevance coexist with low factual accuracy. Across all models, 12 of 14 exceed 94\% on Link Works and all frontier models exceed 80\% on Relevant Content. However, Fact Check scores range from 24\% (OSS-120B) to 77\% (Claude Opus 4.5), a 53\% spread that makes factual accuracy the most differentiating dimension. This pattern means that a user encountering a citation in an LLM-generated report will almost always find a working link to a topically relevant page, yet the specific factual claims attributed to that source may be unsupported nearly half the time.

Open-source models struggle with the citation generation task itself. Fewer than half of open-source models successfully generate cited reports. OSS-120B achieves 40\% task success, Llama 4 Maverick 30\%, and Pixtral Large only 17\%, compared to 83--100\% for frontier models. These models also make substantially fewer tool calls (7--60 versus 73--211 for frontier models), suggesting that limited search capabilities compound with weaker generation abilities.

\subsection{Finding 2. Citation quantity trades off against factual accuracy}
\label{sec:finding2}

Providers that generate more citations achieve lower factual accuracy. Performance patterns differ substantially across providers. OpenAI models achieve 100\% task success and generate the most citations (GPT-5 Mini, 1,272 total attributions), but their Fact Check accuracy spans only 39--59\%. Anthropic models show lower task success (83--97\%) but excel in factual accuracy, with Claude Opus 4.5 achieving 77\% Fact Check and Claude Haiku 4.5 achieving 69\%. Google Gemini models occupy a middle ground (45--49\% Fact Check). This inverse relationship between citation quantity and quality suggests that current LLMs face a fundamental trade-off between research thoroughness and factual reliability.

We hypothesize that this trade-off arises from attention dilution during synthesis. Models generating more citations must aggregate information from a larger number of retrieved passages, increasing the likelihood of misattribution or conflation of facts across sources~\citep{lumer2025memtool}. Notably, Claude Opus 4.5 achieves the highest Fact Check score despite a lower task success rate than any OpenAI model, suggesting that selective citation may be a more effective strategy than exhaustive citation for maintaining factual accuracy.

\subsection{Finding 3. More search degrades factual accuracy}
\label{sec:finding3}

\begin{figure}[t]
\centering
\includegraphics[width=\columnwidth]{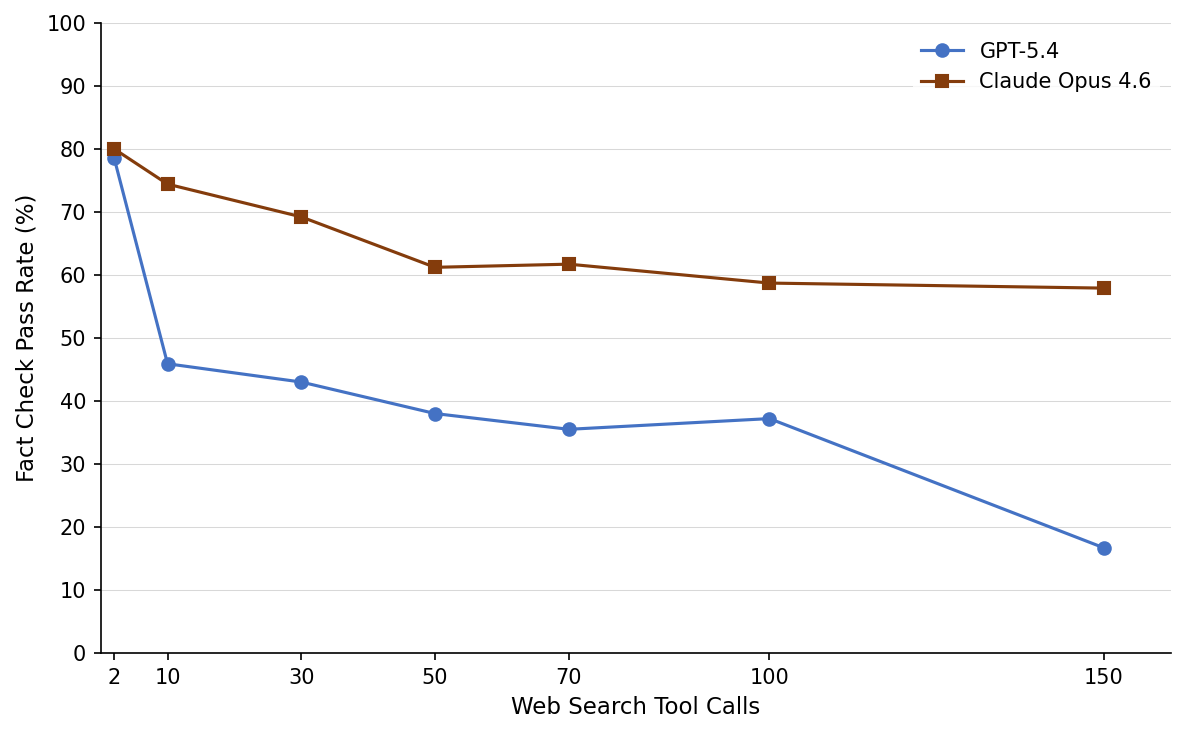}
\caption{Fact Check accuracy degradation as search depth increases. Both models show declining factual accuracy with increased tool calls, while Link Works and Relevant Content remain stable (above 92\%). GPT-5.4 exhibits the steepest decline (79\% to 17\%).}
\label{fig:ablation}
\end{figure}

To directly test the information overload hypothesis, we conducted an ablation study across two models at seven search depth intervals (2--150 tool calls). Tables~\ref{tab:ablation_gpt}--\ref{tab:ablation_opus} and Figure~\ref{fig:ablation} present the results.

\begin{table}[t]
\centering
\caption{GPT-5.4 attribution quality across search depths. \textbf{Bold} indicates best performance.}
\label{tab:ablation_gpt}
\begin{tabular}{lccc}
\toprule
\textbf{Tool Calls} & \textbf{Link Works} & \textbf{Relevant} & \textbf{Fact Check} \\
\midrule
2 & 100.0\% & 100.0\% & \textbf{78.6\%} \\
10 & 100.0\% & 99.0\% & 45.9\% \\
30 & 98.5\% & 97.8\% & 43.0\% \\
50 & 98.6\% & 96.5\% & 38.0\% \\
70 & 100.0\% & 99.1\% & 35.5\% \\
100 & 97.7\% & 95.3\% & 37.2\% \\
150 & 99.2\% & 99.2\% & 16.7\% \\
\bottomrule
\end{tabular}
\end{table}

\begin{table}[t]
\centering
\caption{Claude Opus 4.6 attribution quality across search depths. \textbf{Bold} indicates best performance.}
\label{tab:ablation_opus}
\begin{tabular}{lccc}
\toprule
\textbf{Tool Calls} & \textbf{Link Works} & \textbf{Relevant} & \textbf{Fact Check} \\
\midrule
2 & 100.0\% & 100.0\% & \textbf{80.0\%} \\
10 & 92.3\% & 92.3\% & 74.4\% \\
30 & 100.0\% & 100.0\% & 69.2\% \\
50 & 98.0\% & 98.0\% & 61.2\% \\
70 & 100.0\% & 97.9\% & 61.7\% \\
100 & 100.0\% & 100.0\% & 58.7\% \\
150 & 100.0\% & 100.0\% & 57.9\% \\
\bottomrule
\end{tabular}
\end{table}

Factual accuracy generally degrades with search depth while surface metrics remain stable. Across both models, Fact Check accuracy drops approximately 42\% on average from minimal (2 calls) to maximal search depth (Figure~\ref{fig:ablation}). GPT-5.4 shows the steepest decline, from 79\% to 17\% (62\%). Claude Opus 4.6 demonstrates the greatest resilience, declining from 80\% to 58\% (22\%). Critically, Link Works and Relevant Content remain above 92\% at all search depths, indicating that the degradation is specific to factual synthesis rather than source selection.

This asymmetric degradation pattern provides strong evidence for an information overload effect in LLM research synthesis~\citep{lumer2025scalemcp}. Models can consistently identify and cite accessible, topically relevant sources regardless of search depth, but accurately synthesizing factual claims becomes increasingly difficult as the number of sources grows. The sharpest Fact Check decline occurs between 2 and 10 tool calls for GPT-5.4 (79\% to 46\%, a 33\% drop), suggesting that even modest increases in source volume can overwhelm factual synthesis capabilities.

\subsection{Error analysis}

Link Works failures stem from three categories, including HTTP 404 errors (content removed or URL changed), blocked access (paywalls, bot detection), and connection timeouts. GPT-5.4 demonstrated the highest link reliability with only 1 failed link out of 2,159 evaluations, while open-source models showed substantially higher failure rates (Llama 4 Maverick at 19.2\%, OSS-120B at 16.1\%). Rate limit errors (HTTP 429) had minimal impact across all models, affecting fewer than 0.3\% of evaluations with differences of less than 0.5\% in adjusted pass rates.

\section{Limitations}

Our framework has several limitations that present opportunities for future work. First, the LLM-as-a-judge approach for Relevant Content and Fact Check evaluations, despite calibration through human review, may retain biases inherent to the judge model, including position bias and self-enhancement effects~\citep{wang2023large, zheng2024judging}. Ensemble judging with multiple LLM evaluators or hybrid approaches combining LLM judges with rule-based verification could mitigate single-model biases.

Second, web citations are temporally unstable. URLs that were accessible during evaluation may become unavailable due to content removal, domain expiration, or access policy changes, and source content itself may change. This temporal instability affects reproducibility and motivates longitudinal studies tracking citation persistence over time~\citep{sen2026chronos}. Deploying LLM-generated research in high-stakes domains such as healthcare, legal, or financial analysis without robust citation verification could propagate factual errors, underscoring the importance of frameworks like ours as a safeguard.

Finally, the evaluation is limited to models with web search capabilities, excluding enterprise RAG deployments that cite internal document corpora. Extending the framework to evaluate citation quality in private knowledge bases~\citep{gulati2026beyond} would broaden applicability to the growing ecosystem of enterprise AI assistants.

\section{Conclusion}

In this work, we introduce a source attribution evaluation framework that extracts and evaluates inline citations from LLM-generated deep research reports across three dimensions of link accessibility, topical relevance, and factual accuracy. Our evaluation of 14 LLMs reveals a critical disconnect between surface-level citation quality and factual reliability. Models consistently produce working links to relevant pages, yet factual accuracy remains the weakest dimension across all providers. We further demonstrate an information overload effect, where increased search depth degrades factual accuracy while surface metrics remain stable, providing evidence that more retrieval does not produce more accurate citations.

These findings have implications for both system designers and end users of deep research agents. For system designers, our results suggest that citation quality monitoring should be integrated into agent pipelines, and that retrieval strategies should prioritize depth of source understanding over breadth of source coverage. For end users, the high link validity and topical relevance scores may create a false sense of trust that our framework can help calibrate. We hope that future work will extend this evaluation infrastructure to domain-specific research tasks, enterprise RAG deployments, and longitudinal studies of citation persistence, advancing the systematic assessment of attribution quality as LLM-generated research becomes increasingly prevalent.

\section*{Ethics Statement}

This work evaluates the factual reliability of citations produced by commercially deployed LLM-based deep research agents. Our findings that surface-level citation quality masks factual failures have direct implications for users who rely on these systems for research, decision-making, and information synthesis. We highlight the risk that high link validity and topical relevance scores may create a false sense of trust in LLM-generated citations, particularly in high-stakes domains such as healthcare, legal, and financial analysis where factual errors can cause tangible harm. Our framework is intended to serve as a safeguard by providing transparent, reproducible evaluation of citation quality, and we encourage system designers to integrate similar evaluation mechanisms into deployed research agents. All models evaluated in this study are publicly accessible, and our evaluation methodology does not involve human subjects, personally identifiable information, or sensitive data.

\bibliographystyle{colm2026_conference}
\bibliography{references}

\end{document}